\def\year{2020}\relax
\documentclass[letterpaper]{article} 
\usepackage{aaai20}  
\usepackage{times}  
\usepackage{helvet} 
\usepackage{courier}  
\usepackage[hyphens]{url}  
\usepackage{graphicx} 
\usepackage{bbding}
\usepackage{tabularx}
\usepackage{multirow}
\usepackage{setspace}
\usepackage{color,subfigure}
\usepackage{mathrsfs}
\usepackage{graphics}
\usepackage{epsfig}
\usepackage{epstopdf}
\usepackage{url}
\usepackage{graphicx}
\usepackage{array}
\usepackage{times}
\usepackage{epsfig}
\usepackage{graphicx}
\usepackage{amsmath}
\usepackage{amssymb}
\usepackage{graphicx}  
\title{Identifying and Resisting Adversarial Videos Using Temporal Consistency }
\author{Xiaojun Jia$^1$$^2$, Xingxing Wei$^1$, Xiaochun Cao$^1$$^2$\\ \textsuperscript{\rm 1}Institute of Information Engineering, Chinese Academy of Sciences \\ \textsuperscript{\rm 2}Cyberspace Security Research Center, Peng Cheng Laboratory, Shenzhen 518055, China\\ \textsuperscript{\rm 3}Beihang University}
 
\begin{document}

\maketitle

\begin{abstract}
Video classification is a challenging task in computer
vision. Although Deep Neural Networks (DNNs) have
achieved excellent performance in video classification, recent research shows adding imperceptible perturbations
to clean videos can make the well-trained models output
wrong labels with high confidence. In this paper, we propose an effective defense framework to characterize and
defend adversarial videos. The proposed method contains two phases: (1) adversarial video detection using \emph{temporal consistency} between adjacent frames, and (2) adversarial perturbation reduction via denoisers in the spatial
and temporal domains respectively. Specifically, because of
the linear nature of DNNs, the imperceptible perturbations will enlarge with the increasing of DNNs' depth, which
leads to the inconsistency of DNNs' output between adjacent frames. However, the benign video frames often have
the same outputs with their neighbor frames owing to the
slight changes. Based on this observation, we can distinguish between adversarial videos and benign videos. After
that, we utilize different defense strategies against different attacks. We propose the \emph{temporal defense}, which reconstructs the polluted frames with their temporally neighbor clean frames, to deal with the adversarial videos with
sparse polluted frames. For the videos with dense polluted
frames, we use an efficient adversarial denoiser to process
each frame in the spatial domain, and thus purify the perturbations (we call it as \emph{spatial defense}). A series of experiments conducted on the UCF-101 dataset demonstrate that the proposed method significantly improves the robustness of video
classifiers against adversarial attacks.

\end{abstract}

\begin{figure*}[tt]
\begin{center}
   \includegraphics[width=0.8\linewidth]{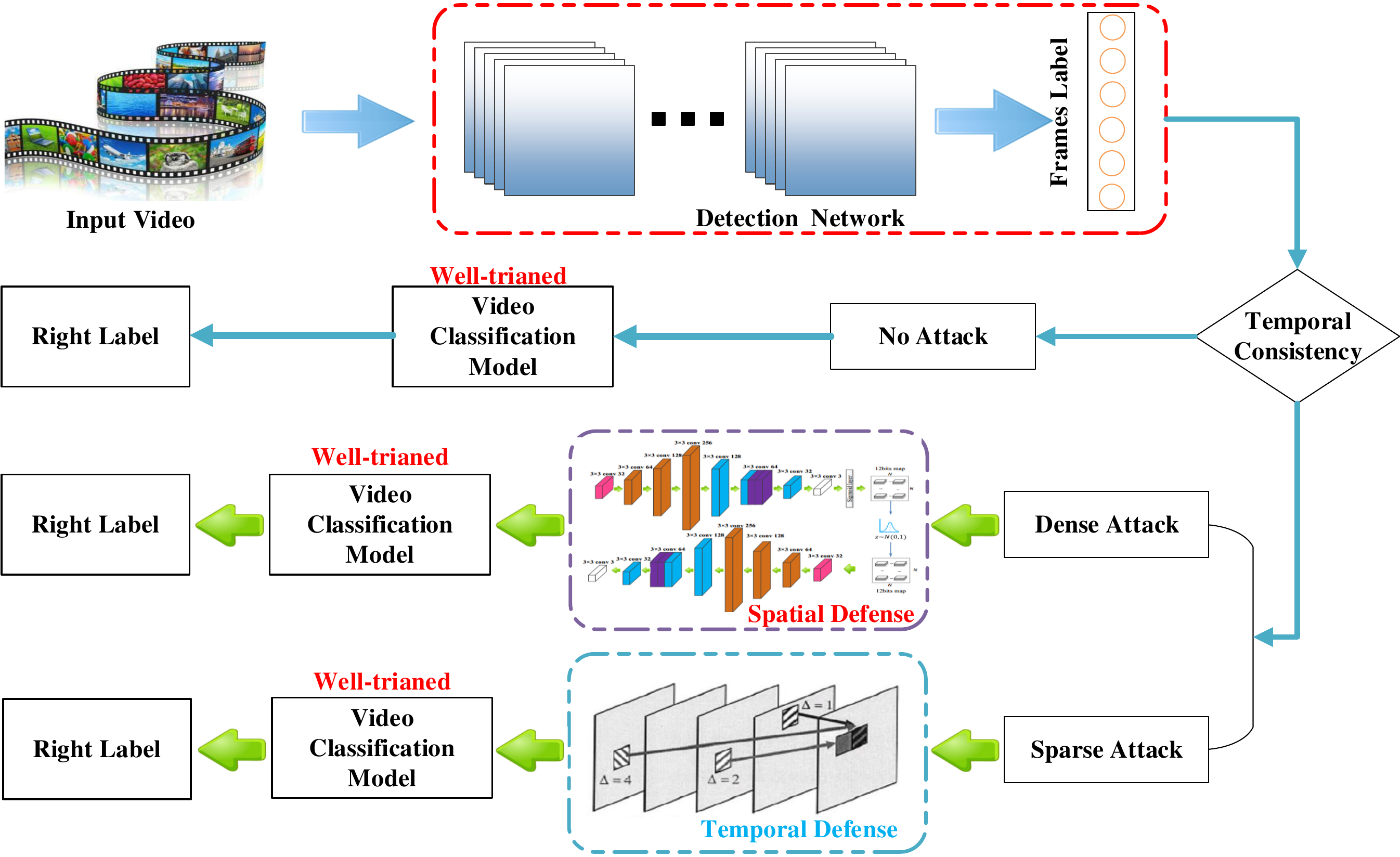}
\end{center}
   \caption{The proposed framework consists of two steps: adversarial video detection and  adversarial video defense.
   In the detection stage, the temporal consistency of videos is used to achieve the goal.
   In particular, we classify the video into three types (clean videos, sparse adversarial videos and dense
   adversarial videos) by quantitative analysis of the detection network outputs. In the defense stage,
   our method consists of temporal defense and spatial defense. The spatial defense
   is used to defend the dense adversarial attacks. And the temporal defense
   takes charge of defending the sparse adversarial attacks.
   }
\label{fig:iccv01}
\end{figure*}

\section{Introduction}
With the rapid growth of video data on the internet,  automatic video classification \cite{karpathy2014large} is becoming more and more important. Specifically, this task includes action recognition \cite{yue2015beyond}, scene classification \cite{huang1999integration},
object classification \cite{zhang2007real} and so on. Inspired by the great performance of Convolutional Neural
Networks (CNNs)\cite{lecun2015deep}, more and more researchers have begun to apply the deep
learning to the video classification task \cite{Karpathy_2014_CVPR}, and achieved great success. However, some research works \cite{wei2018sparse,li2018adversarial,zajkac2018adversarial}  show  that video classification is vulnerable to the adversarial examples \cite{nguyen2015deep}, just like the case in images \cite{szegedy2013intriguing,goodfellow2014explaining}.
Different from static images, attacking video classification not only needs to consider the spatial information but also needs to consider the temporal information.

The study for video attacks is less so far. To our knowledge, there are only three related papers \cite{wei2018sparse,li2018adversarial,zajkac2018adversarial}. In summary, these  methods can be roughly divided into two classes, the first one can be called as \textbf{sparse attack}, which denotes only
 some key frames in a video are polluted, the generated adversarial perturbations are temporally sparse. For example, in \cite{wei2018sparse},
 the authors propose that the slight perturbations added the current frame can transfer to the next
frames via the temporal interactions between frames, and thus don't need to pollute every frame.  Using iterative optimization algorithm
based on \begin{math}l_{2,1}-\end{math}norm, they successfully attack the state-of-the-art video classification model.
The other one can be called as  \textbf{dense attack}, which means that all the frames in a video are polluted, such as in \cite{li2018adversarial}, a generative method for adversarial perturbations  is proposed to attack the real-time video classification system. Because the test phase only involves a feedforward network, the attacking efficiency is high. Besides, it also explores a robust 3D adversarial perturbations  to overcome the varying boundaries of video clips.

To defend adversarial examples, many kinds of research techniques have been proposed. In the image case,
 adversarial training \cite{tramer2017ensemble} is a widely used technique, it improves the DNNs' robustness by adding adversarial examples into the
 training dataset to retrain the classification model.  In addition, \cite{liao2018defense} proposes that
image denoising can be used to perform the defense task, and designs a High-level representation Guided Denoiser (HGD) to remove the adversarial
perturbations. Inspired by this, ComDefend \cite{jia2018comdefend} argues that  image compression is useful to defend adversarial examples,
and presents an end-to-end image compression model to achieve defense. We can see all the above methods are designed for static images classification. The temporal information within videos are not considered,  therefore, these methods are not completely suitable to defend the adversarial videos. We need a new framework to defend adversarial examples for video classification.

For this reason, in this paper, we propose an effective defense framework to characterize and defend video adversarial examples.
 Our method contains two steps. The first step is to detect the adversarial videos using temporal consistency between adjacent frames.  If the input video is benign, we directly feed it to the well-trained video classifier. Otherwise, an extra pre-processing module is used to denoise it.
 We accomplish this task by using the temporal consistency \cite{xiao2018monet}(video frames which are temporally close have similar image characteristics).
 It is known that one accepted reason behind adversarial
 examples is the linear nature of DNNs \cite{goodfellow2014explaining}.  The imperceptible perturbations added on the images will enlarges with the increasing of DNNs' depth,  which leads to the inconsistency of DNNs' output between adjacent frames. By contrast, the benign video frames often have the same outputs with their neighbor frames owing to the slight changes. This difference can help us distinguish between adversarial videos and benign videos. In the implementation, we present a metric to represent the degree of temporal consistency, and then use a threshold to classify them. The second step is to reduce the adversarial perturbations via the different denoisers in the spatial and temporal domains respectively.
 As for the sparse attack \cite{wei2018sparse}, we propose the \emph{temporal defense} method, which utilizes the temporal interactions between frames,
 and reconstructs the polluted frames with their temporally neighbor clean frames.
 As for the dense attack \cite{li2018adversarial}, we use the \emph{spatial defense} method,
 which uses an efficient adversarial denoiser to process each frame in the spatial domain, and obtain their clean versions.
 We propose the corresponding defense strategies according to the properties of the attacks. Experiments show that temporal defense obtains the best performance against sparse attack, and spatial defense is also the best to defend the dense attack.
Figure \ref{fig:iccv01} illustrates the overall framework.

\par In summary, this paper has the following contributions:
\par (1) To the best of our knowledge, we are the first one to explore the defense for  adversarial videos, and further, propose a two-step defense
framework for video classification. The proposed framework utilizes different defense strategies according to the properties of attacks.
Experiments show that our method significantly improves the robustness of video classifiers.
\par (2) We design the adversarial video detection method based on the temporal consistency between adjacent frames.  We find that the linear nature of DNNs will lead to the temporal inconsistency for adversarial videos, and further, present a metric to represent the degree of temporal consistency.  Using this metric, a simple threshold can be used to classify the adversarial videos.
\par (3) We propose the temporal defense against the sparse attack. Temporal defense utilizes the temporal interactions between adjacent frames, and reconstructs the polluted frames with their neighbor clean frames, and thus  purifies the adversarial perturbations.  Temporal defense is the first one to explore the structured information in videos to perform adversarial defense task.

\par The remainder of this paper is organized as follows. Section 2 briefly reviews the related work. Section 3 introduces the details of the proposed framework. Section 4 shows a series of experimental results and analysis. Finally, Section 5 gives the
conclusion.
\section{Related work}

\subsection{Video classification}
Currently, there are three kinds of deep learning methods for video classification. The first one is to use the existing
image classification methods to achieve video classification. In particular, this method regards the
video as a collection of frames. Typically, it uses the pre-trained network on
ImageNet \cite{deng2009imagenet} to extract frame features and then overlays them
as video features to perform classification. The second one is to use an end-to-end 3D CNN architecture
to achieve video classification. The difference with the first method lies in that the part of feature extraction is
directly trained on the video data, such as \cite{ji20133d,tran2015learning}. The third one is to
use CNN+LSTM to achieve video classification.
These methods often use LSTM \cite{hochreiter1997long} to extract temporal information of videos.
 A CNN is firstly used to extract the frame feature, and then a LSTM is used to explore the temporal relationship between frames, such as \cite{donahue2015long,yue2015beyond}.
 In addition, a kind of two-stream architecture \cite{simonyan2014two} and its variants \cite{feichtenhofer2016convolutional,wu2015modeling}, where one stream is to encode the spatial information within frames, and another stream is to encode the temporal information (usually optical flow is extracted to represent the temporal information) between frames,  are proposed and achieve the competitive performance.

\subsection{Adversarial attacks}
Besides the image classification, video classification models are also vulnerable to adversarial examples.
\cite{wei2018sparse} proposes a kind of sparse attacking method to attack
the CNN+RNN architecture, which is widely used in the video classification task. Their research demonstrates
that the adversarial perturbations can be transferred between the video frames. Simultaneously,
a dense adversarial attack method is proposed in \cite{li2018adversarial}. It attacks the C3D
model \cite{tran2015learning}, which is also a widely used method in video classification. The attacking method in
\cite{li2018adversarial} is based on a
generating model. And each frame of the video is
added with the generated adversarial perturbations. Also, an adversarial framing is proposed
in \cite{zajkac2018adversarial}. Compared with \cite{tran2015learning}, this method does not
modify the most pixels of the input image. It just adds an adversarial framing on the image
border.

\subsection{Adversarial defense}
In \cite{tramer2017ensemble}, F. Tramer \emph{et al.} propose the adversarial training method to
retrain the classification model by adding the adversarial examples  into the original training dataset. The used adversarial examples are pre-generated by different
attack methods. In \cite{liao2018defense}, the adversarial perturbations
are regarded as a special kind of image noise. HGD is a denoising model for dealing with such noise.
It is also trained by the generated adversarial images. And in \cite{jia2018comdefend},
an end-to-end compression model is proposed to defend the adversarial examples. It uses image
compression to break the structure of the adversarial perturbations for defense. In addition, there are many other defense methods, like \cite{pang2018max,xie2017mitigating,athalye2018obfuscated}, and so on. The above defense methods are all focused on the images. They only consider the spatial information, but ignore the temporal information. Therefore, these methods are not completely suitable for the video classification task.

\section{Methodology}
\subsection{The framework of the proposed method}
According to the number of polluted frames in the video data, the current video attacking methods
can be divided into sparse attacks and dense attacks. The different attacks are defended by the corresponding defense
methods in our paper. The proposed framework consists of two phases. The first one is the detection network which
is used to determine whether the video is attacked, and further, make sure of the type of the attack. And then,
different defense methods are carried out according to the detection result. As for the sparse adversarial
video attack, which modifies the fewer frames of the video, we use the temporal defense to transform the adversarial
video into the corresponding clean video. As for the dense adversarial video attack, which modifies all the frames of the video,
we use the spatial defense to transform the adversarial video into the corresponding clean video.

\begin{figure}[tt]
\begin{center}
   \includegraphics[width=0.9\linewidth]{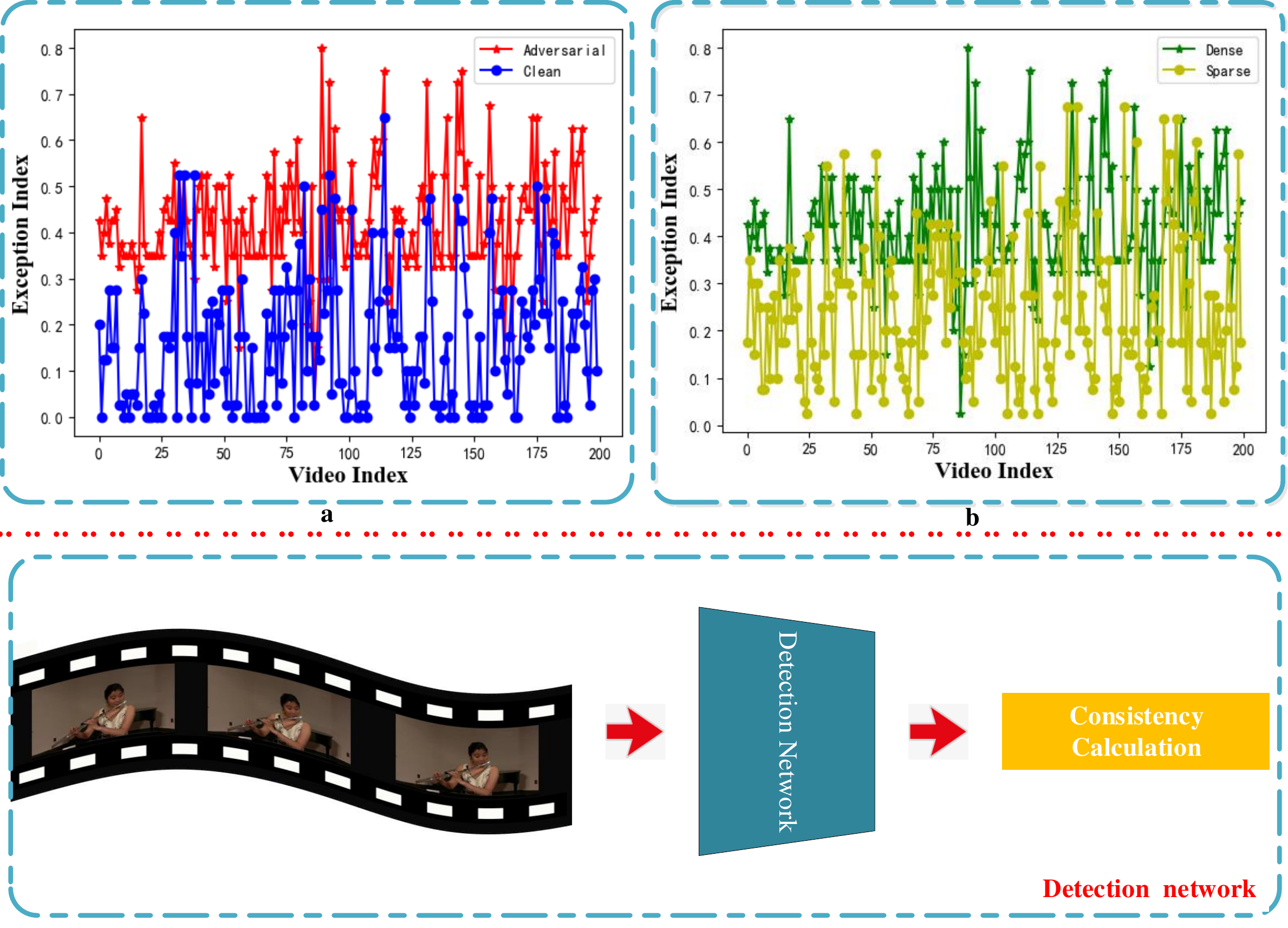}
\end{center}
   \caption{The detection stage includes the detection network and consistency
   calculation. We randomly select 400 video samples
   which consist of 200 clean videos and 200 adversarial videos, and then compute their exception index $\alpha$. The results are shown in
   the sub-figure (a). The red line represents the results of adversarial videos, and the blue line represents the results of clean videos. In addition, we also randomly
   select 400 video samples which consist of 200 sparse adversarial
   videos and 200 dense adversarial videos. The computed exception index $\alpha$ values are shown in
   the sub-figure (b). The green line represents the dense adversarial videos, and the yellow line represents the sparse adversarial videos. We can see that the $\alpha$ values of clean videos, sparse adversarial videos, and dense adversarial videos are linearly classified. We can use the simple thresholds to classify them.
   }
\label{fig:iccv02}
\end{figure}

\begin{figure*}[tt]
\begin{center}
   \includegraphics[width=0.9\linewidth]{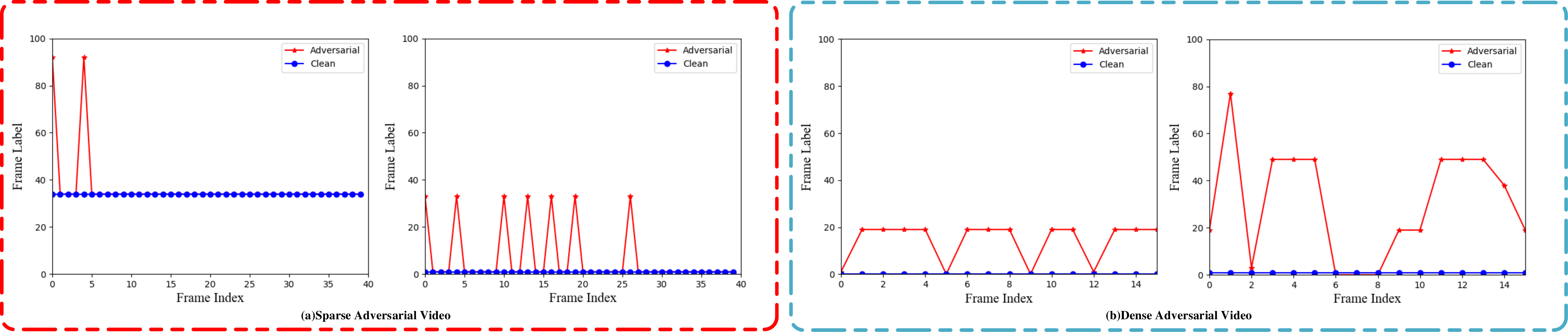}
\end{center}
   \caption{The left column is the label of the sparse adversarial video frames and their
   corresponding clean video frames. The right column is the label of
   the dense adversarial video frames and their corresponding clean video frames.
   }
\label{fig:consistency}
\end{figure*}

\subsection{Detection network}
\begin{figure}[tt]
\begin{center}
   \includegraphics[width=0.9\linewidth]{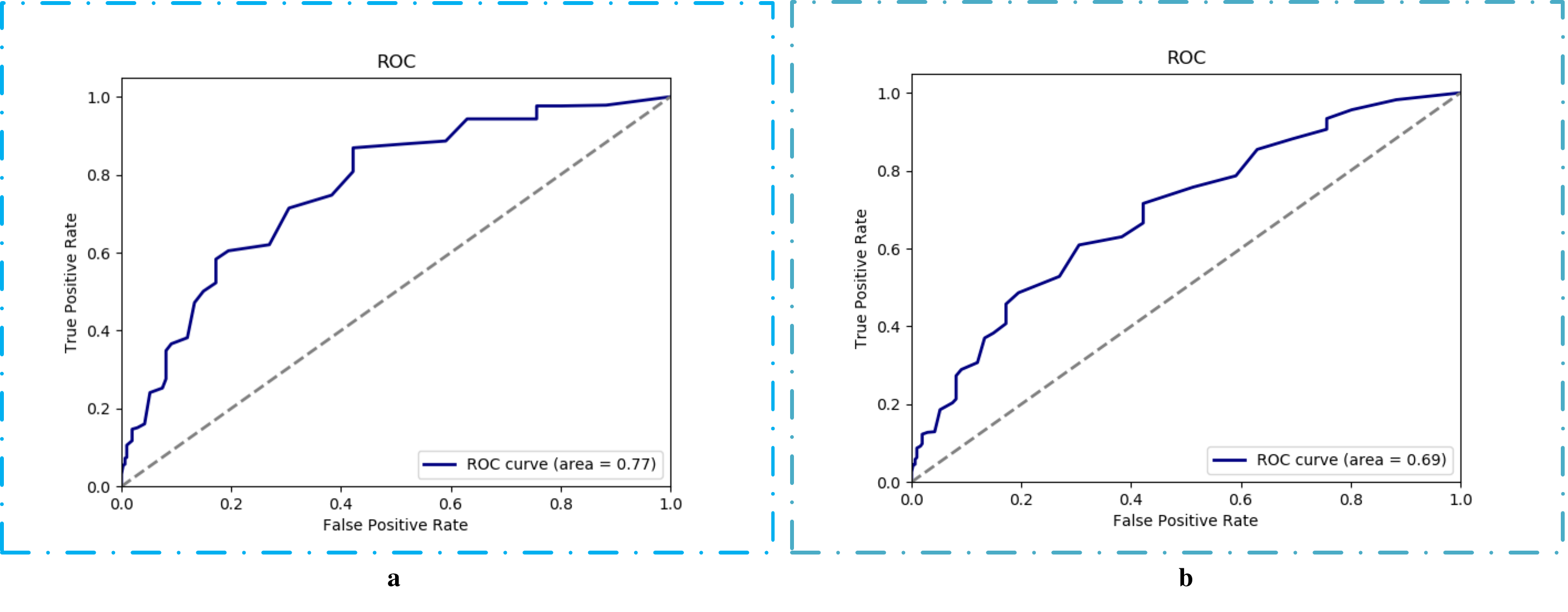}
\end{center}
   \caption{The ROC curves of the proposed detection method under different thresholds.
   Subfigure (a) represents the detection performance on the clean and adversarial videos.
   Subfigure (b) represents the detection performance  on the sparse and dense adversarial videos.
   }
\label{fig:iccvROC}
\vspace{-.2cm}
\end{figure}

There are two main functions of the detection network: (1) detecting whether the video is attacked, and (2) detecting what
kind of attack the video is subjected to. There is a high correlation between the adjacent frames of the video.
In \cite{xiao2018monet}, in the video object segmentation task, they propose that the video frames
which are temporally close have similar image characteristics. This phenomenon can be called as the
temporal consistency of the video data. Through a series of experiments, we find that this phenomenon also exists
in the video classification task. In particular, the outputs produced by the well-trained video classification
maintain consistency in the adjacent frames of the benign video. By contrast, because of the linear nature of DNNs, the imperceptible adversarial perturbations  will enlarge with the increasing of DNNs' depth,  which leads to the inconsistency of DNNs' outputs between the adjacent frames in the adversarial video. As shown in Figure \ref{fig:iccv02}, we feed each
frame of the input videos to the well-trained network for video classification to get the classification label.
The CNN+LSTM architecture \cite{Donahue_2015_CVPR} is used as the well-trained network in this paper.
Note that in order to eliminate the inter-frame effects, we remove the links between LSTMs in the
detection network. If the label of the current frame is different from the labels of its adjacent frames,
the current frame is regarded as an exception frame. We define an exception index \begin{math} \alpha \end{math} to represent the degree of temporal consistency. It is defined as follows:
\begin{equation}\label{mvr}
\begin{split}
\alpha=\frac{1}{T}\sum_{n=2}^{T} [f(n-1)\neq f(n) \wedge f(n+1)\neq f(n)],
\end{split}
\end{equation}
where \begin{math} f(n) \end{math} represents the predicted label of the $ n$-th frame. And \begin{math} T \end{math}
represents the total number of frames in a video. According to the value of \begin{math} \alpha \end{math}, we
can perform the detection. If \begin{math} \alpha <\gamma_{1} \end{math}, the video is benign.
If \begin{math} \gamma_{1}\leq\alpha <\gamma_{2} \end{math}, the video is subjected to the sparse attack.
And if \begin{math} \gamma_{2}\leq\alpha  \end{math}, the video is subjected to the dense attack.
In addition, \begin{math} \gamma_{1} \end{math} and \begin{math} \gamma_{2} \end{math} are hyper-parameters
which are determined in the experiments. We use the Receiver Operating Characteristic (ROC) curve to
evaluate the proposed detection method. As shown in Figure \ref{fig:iccvROC}, the proposed method is appropriate
to detect whether the original video is attacked and determine the type of the attack. We use the proposed detection
network to deal with randomly selected two sparse adversarial videos and dense adversarial videos. The result is shown
in Figure \ref{fig:consistency}.

\subsection{Spatial defense}
\begin{figure}[tt]
\begin{center}
   \includegraphics[width=0.9\linewidth]{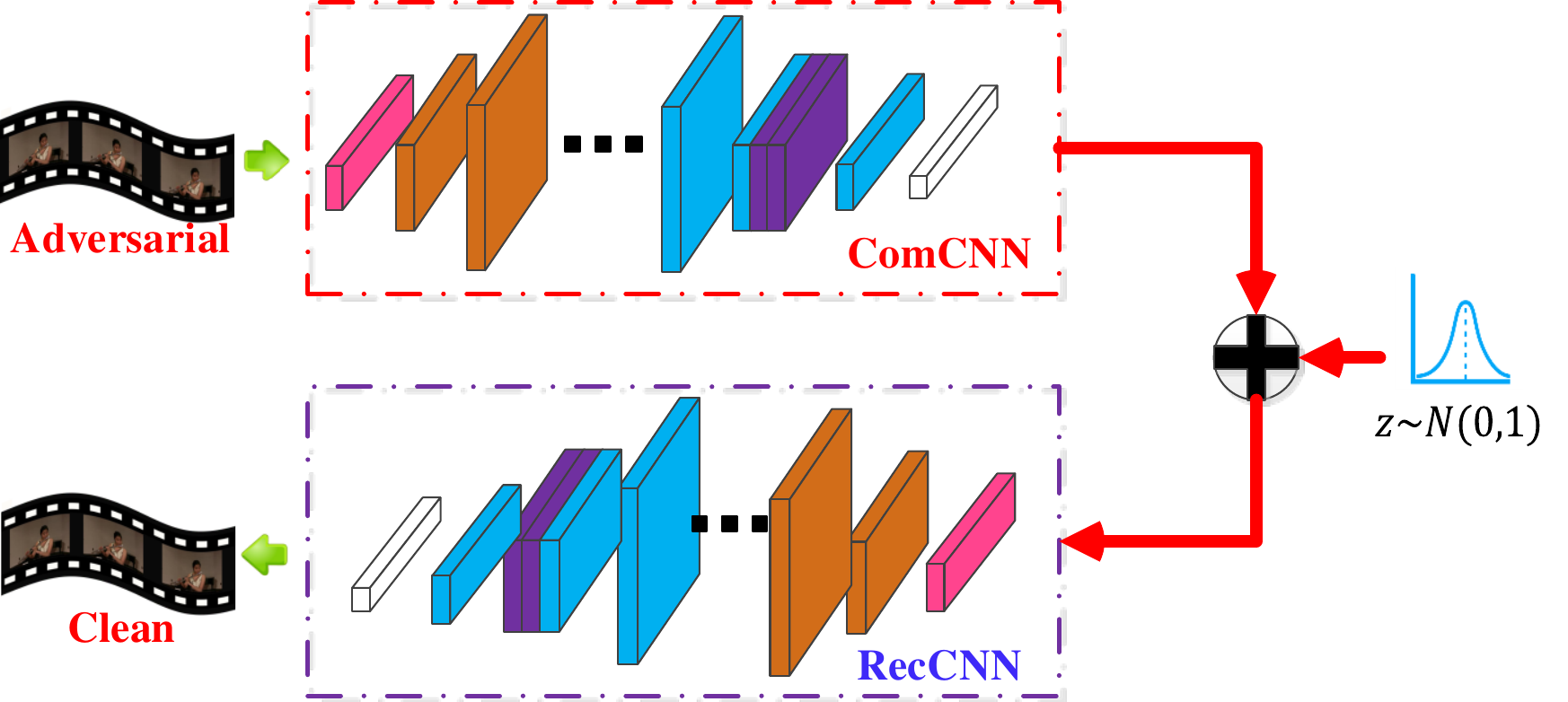}
\end{center}
   \caption{The spatial defense is achieved using the ComDefend  \cite{jia2018comdefend}, which includes the ComCNN and RecCNN modules.
   Each frame of the video is fed into the ComCNN to achieve the spatial compression.
   And the RecCNN is used to reconstruct each frame of the video. We use the ComDefend
   to remove spatial redundancy and adversarial perturbations. Because ComDefend is very efficient, it is quick to deal with the whole video.
   }
\label{fig:iccv03}
\end{figure}

In order to transform the dense adversarial videos to the corresponding clean videos, we use an efficient denoiser to process each frame. We here select an end-to-end compression model.
As shown in Figure \ref{fig:iccv03}, the whole process includes two modules: image compression module and image reconstruction module. During
the image compression stage, the ComCNN is used to compress each frame and extract its main structure
information. During the image reconstruction stage, the RecCNN uses the compressed information of each frame
to reconstruct the corresponding clean frame. For more details, please refer to \cite{jia2018comdefend}. Because the operation is conducted within the spatial domain of each frame, this method is called as \textbf{spatial defense}.

\subsection{Temporal defense}
There is a high correlation between the temporally adjacent frames of the video. The adversarial
frames can be replaced by the pseudo frames which are reconstructed by the adjacent clean frames. Therefore,
we propose to use the motion estimation method to compress the sparse adversarial videos for defense. In this way,
the sparse adversarial video can be transformed into the corresponding clean video.

\par Motion estimation \cite{furht2012motion} is a widely used technique in video processing. The motion
estimation can be achieved based on image patch or image grid. Patch-based motion estimation is widely used
because of its simple algorithm and easy implementation. We still focus on it. The basic idea of motion
estimation is to divide each frame of videos into a number of non-overlapping patches, and consider
that all pixels in the patches have the same amount of displacement, and then find the patch that is most
similar to the current patch in the reference frame according to certain matching criteria, called matching patch.
The relative displacement of the matching patch and the current patch is the motion vector.
When the video is compressed, we only need to save the motion vector and corresponding residual data
to recover the current patch. Note that using integer Discrete Cosine Transform (DCT) \cite{ahmed1974discrete}
to compress the corresponding residual data.
The motion vector is defined as:
\begin{equation}\label{mvr}
\begin{split}
{\rm MAD}_{(p,q)}(i,j)=\frac{1}{M \ast N}\sum_{m=0}^M\sum_{n=0}^N|f_{k}(p+m,q+n) \\
-f_{k-1}(p+m+i,q+n+j)| ,
\end{split}
\end{equation}
where \begin{math}f_{k}(m,n)\end{math} represents the intensity of the pixel with
coordinates \begin{math}(m,n)\end{math} in the \begin{math}{\rm k}th\end{math} frame.
\begin{math}M\end{math} represents the width of the patch. \begin{math}N\end{math} represents the length of the patch.
We use the patch's upper left corner coordinates
\begin{math}(p,q)\end{math} to represent it. And \begin{math}\rm MAD\end{math}
represents the similarity between the present frame patch \begin{math}(p,q)\end{math} and the previous frame patch \begin{math}(p+i,q+j)\end{math}.
\begin{equation}\label{mvr}
   {\rm Vec}(i,j)= \arg\min {\rm MAD}_{(p,q)}(i,j),
\end{equation}
where \begin{math}{\rm Vec}(i,j)\end{math} represents the motion vector
of the present frame patch \begin{math}(p,q)\end{math}.

\par
\begin{figure}[tt]
\begin{center}
   \includegraphics[width=1.0\linewidth]{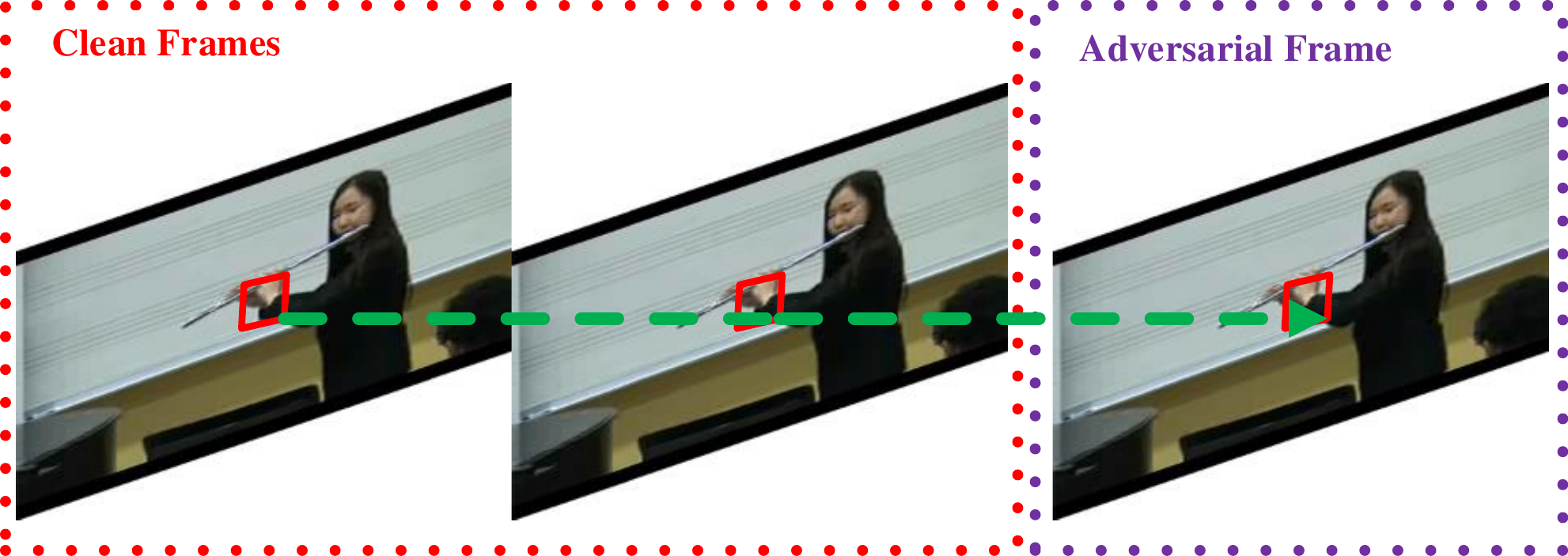}
\end{center}
   \caption{The adversarial video is reconstructed by using the information of the
   previous clean videos in the patch level.
   }
   \vspace{-.3cm}
\label{fig:iccv04}
\end{figure}

  There is a high correlation between the contents in the adjacent frames of the video.
As for the sparse adversarial video, a small number of frames  are attacked.
Inspired by the idea of motion estimation, we can use the motion estimation method
to replace the patches of adversarial video frames with the patches of clean video
frames. As shown in Figure \ref{fig:iccv04}, we can save the patches of clean video frames, the
motion vector and the corresponding residual to reconstruct the patches of the
adversarial video frames. The adversarial video frames can be reconstructed
by the clean frame patches. The quantification
of the corresponding residual can remove the adversarial perturbations.
Because the operation is conducted within the temporal domain of the video, this method is called as \textbf{temporal defense}.

\section{Experiments and analysis}
In this section, we conduct a series of experiments to verify the effectiveness of the proposed
framework, which includes: adversarial video generation, threshold selection in the framework,
video classification with the proposed framework and result analysis.
\begin{table*}[t]
\centering
\footnotesize{
\caption{ The selection of parameter $ \gamma_{1}$. }
 \label{table:crossstreetmAPresult1}
\begin{tabular}{|c|c|c|c|c|c|c|c|c|}
\hline
                  & $\gamma_{1} =0.075$ & $\gamma_{1} =0.10$ & $ \gamma_{1} =0.125$ & $\gamma_{1} =0.15$ & $ \gamma_{1} =0.175$ & $ \gamma_{1} =0.20$ & $\gamma_{1} =0.225$ & $ \gamma_{1} =0.25$ \\ \hline
Precision & 58\%   & 60\%    & 60\%   & 63\%    & 68\%   & 66\%    & 70\%   & 70\%    \\ \hline
Recall    & 95\%   & 95\%    & 90\%   & 89\%    & 88\%   & 75\%    & 71\%   & 62\%    \\ \hline
F1-measure & 72\%   & 74\%    & 73\%   & 74\%    & \textbf{77\%}   & 71\%    & 71\%   & 66\%    \\ \hline
\end{tabular}
}
\vspace{-.4cm}
\end{table*}

\begin{table*}[t]
\centering
\footnotesize{
\caption{ The selection of parameter $ \gamma_{2}$. }
 \label{table:crossstreetmAPresult11}
\begin{tabular}{|c|c|c|c|c|c|c|c|c|}
\hline
                        & $\gamma_{2} =0.25$ & $\gamma_{2} =0.275$ & $ \gamma_{2} =0.30$ & $\gamma_{2} =0.325$ & $ \gamma_{2} =0.35$ & $ \gamma_{2} =0.375$ & $\gamma_{2} =0.40$ & $ \gamma_{2} =0.425$ \\ \hline
Precision & 54\%   & 55\%    & 58\%   & 56\%    & 60\%   & 61\%    & 65\%   & 69\%    \\ \hline
Recall    & 75\%   & 75\%    & 75\%   & 63\%    & 63\%   & 63\%    & 52\%   & 52\%    \\ \hline
F1-measure & 62\%   & 64\%    & \textbf{65\%}   & 59\%    & 61\%   & 56\%    & 57\%   & 58\%    \\ \hline
\end{tabular}
}
\vspace{-.2cm}
\end{table*}

\subsection{Datasets}
The detection network is trained on the video dataset UCF101 \cite{soomro2012ucf101} which is
widely used in action recognition. It consists of 13320 videos from 101 action categories.
The action categories of UCF101 include sports, playing musical instruments,
body-motion, human-human interaction and human-object interaction.
In order to ensure the fairness of the experiment, attack methods and defense
method are all conducted on the UCF101. We choose more than 8000 videos for training the video classifier and 3000 videos for testing it. In addition,
As for training the detection network, we also choose these 8000 videos
of UCF101. The original ComDefend is  trained on the CIFAR-10 dataset, we don't modify it and directly use the well-trained model.

\subsection{Adversarial video generation}
As mentioned previously, the video attacks can be divided into the sparse adversarial attacks and
dense adversarial attacks based on the number of polluted frames.
We choose two typical attacking methods: \cite{wei2018sparse} for the sparse attack and
\cite{li2018adversarial} for the dense attack. \cite{wei2018sparse} chooses CNN+LSTM architecture
as threat models. It uses the characteristic of adversarial perturbations
that can propagate between frames to attack the CNN+LSTM architecture.
Because the adversarial perturbations are added on a few video frames, this kind
of attack method belongs to the sparse adversarial attack method.
\cite{li2018adversarial} makes use of a generative model which looks like a Generative Adversarial Network
(GAN) architecture to generate a series of adversarial perturbations. The generated perturbations
are added on each frame of the original video to fool the C3D model. Because all the frames are polluted,  this kind of attack method belongs to dense adversarial attack method. We use these two methods to generate the adversarial videos on the UCF101 dataset, and then perform our framework to defend them.

\subsection{Threshold selection in the detection stage}
In the detection stage, two threshold hyperparameters
\begin{math} \gamma_{1} \end{math} and \begin{math} \gamma_{2} \end{math}
need to be determined by the experiments.
The hyperparameter \begin{math} \gamma_{1} \end{math} is used to
detect whether the input video is attacked. And the hyperparameter
\begin{math} \gamma_{2} \end{math} is used to determine the type of attacks.  To determine  \begin{math} \gamma_{1} \end{math}, we randomly select 500 clean videos from UCF101 dataset, and then generate their corresponding sparse and dense adversarial videos. Thus, we obtain 1500 videos. This is a two-category classification task. We regard the adversarial videos as the positive samples, and
calculate the precision, recall and F1-measure under the different values of the
parameter \begin{math} \gamma_{1} \end{math}.  The results are shown in Table \ref{table:crossstreetmAPresult1}.  The optimal \begin{math} \gamma_{1} \end{math} is selected according to the highest F1 metric, where the threshold simultaneously shows good precision and recall.  Under this setting, the \begin{math} \gamma_{1} \end{math} is set to 0.175.
To determine  \begin{math} \gamma_{2} \end{math}, we only use the 500 sparse adversarial videos and 500 dense adversarial videos to construct a dataset to select the optimal \begin{math} \gamma_{2} \end{math} value.  Like \begin{math} \gamma_{1} \end{math}, we also compute the precision, recall and F1-measure on this dataset. The difference is that dense adversarial videos are regarded as the positive samples in this case. The final results are shown in Table \ref{table:crossstreetmAPresult11}.  We see that the F1-measure achieves the best performance when \begin{math} \gamma_{2}=0.3 \end{math}. Therefore, in the following experiments, we set \begin{math} \gamma_{1}=0.175 \end{math} and \begin{math} \gamma_{2}=0.3 \end{math}.

\subsection{Video defense with the proposed method}
The experiments in this section include (1) video classification with the
spatial defense separately, (2) video classification with the temporal
defense separately, and (3) video classification with the different defense
strategies combined with detection. In this way, it is easy to
see the performance of each module of the proposed method.

\begin{table*}[t]
\centering
\footnotesize{
\caption{ Results of different defense methods }
\label{table:crossstreetmAPresult2}
\begin{tabular}{|c|c|c|c|c|}
\hline
Network                   &Attacks          & No defense & Spatial defense  & Temporal defense \\ \hline
\multirow{2}{*}{CNN+LSTM} & Clean  videos   & 78\%       & 66\%    & 76\%          \\ \cline{2-5}
                          & Sparse \cite{wei2018sparse} & 31\%       & 55\%       & 64\%      \\ \hline
\multirow{2}{*}{C3D}      & Clean  videos  & 89\%       & 81\%     & 88\%        \\ \cline{2-5}
                          & Dense \cite{li2018adversarial} & 59\%       & 73\%         & 59\%     \\ \hline
\end{tabular}
}
\vspace{-.4cm}
\end{table*}

\begin{table*}[t]
\centering
\caption{The ablation study of our method}
\label{table:crossstreetmAPresult3}
\begin{tabular}{|c|c|c|c|c|}
\hline
Defense method  & No defense & Spatial defense only & Temporal defense only & Detection+temporal/spatial defense \\ \hline
Accuracy & 26\%       & 44\%            & 41\%             & \textbf{48\%}              \\ \hline
\end{tabular}
\vspace{-.5cm}
\end{table*}

\begin{figure}
\begin{center}
   \includegraphics[width=0.9\linewidth]{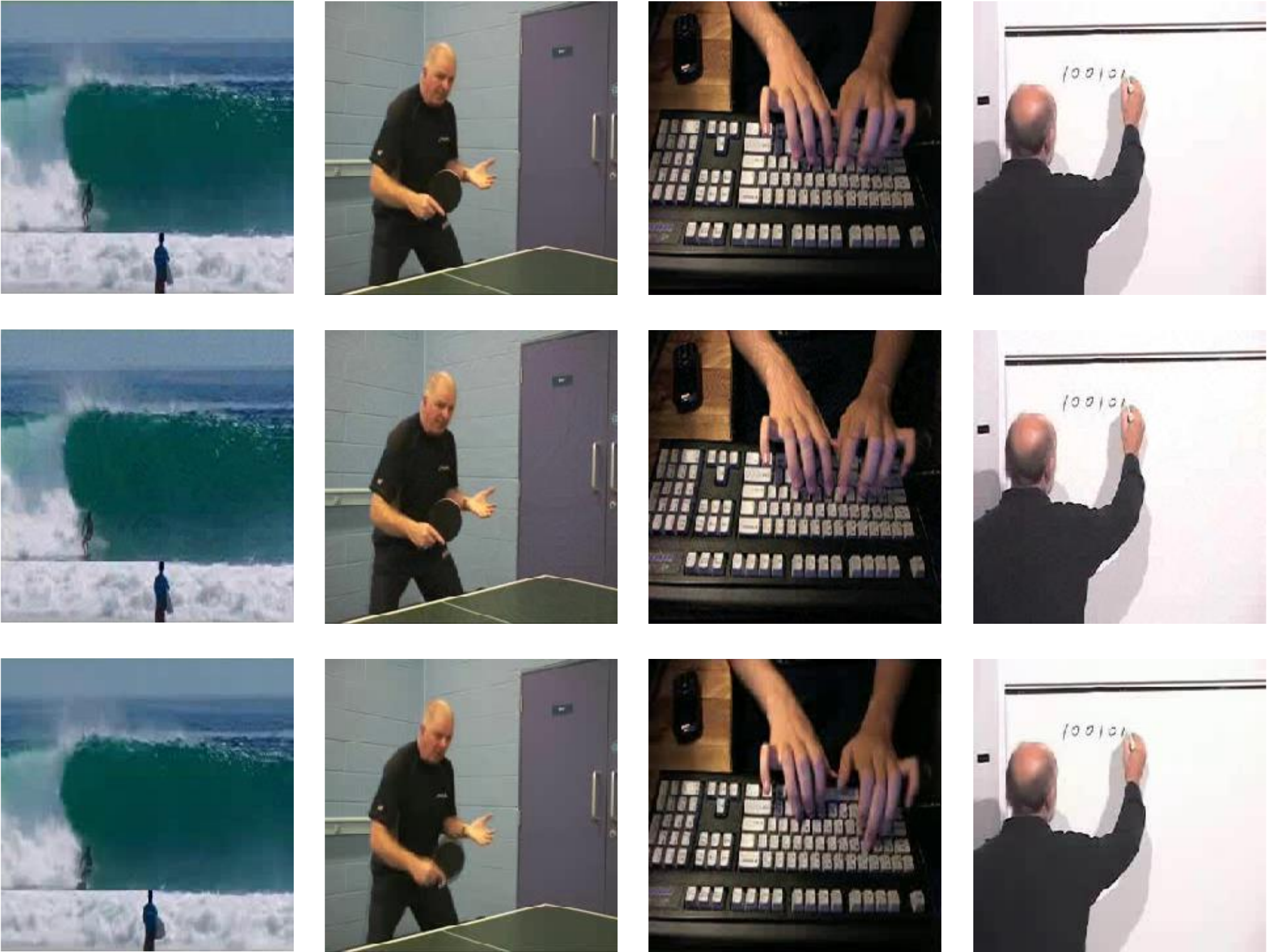}
\end{center}
   \caption{Four examples of the temporal defense for the sparse adversarial attack.
   The top row is the clean video frames, the middle row is the sparse adversarial video frames,
   and the bottom row is the temporal defense video frames.
   }
\label{fig:defense}
\vspace{-.2cm}
\end{figure}
\subsubsection{Video classification with the temporal defense}

The proposed temporal defense is a pre-processing module, and does not modify the deployed model.  When the input video is detected as the sparse attack, we firstly use the temporal defense to denoise it, and then feed it to the deployed model. To evaluate the performance of the temporal defense, we give the experiments in Table \ref{table:crossstreetmAPresult2}. Two deployed video classification models are tested: CNN+LSTM and C3D. We use \cite{wei2018sparse} and \cite{li2018adversarial} to generate the adversarial videos, and the temporal defense
to defend them. In Table \ref{table:crossstreetmAPresult2}, we can see that the temporal defense significantly improves the classification accuracy of CNN+LSTM from 31\% to 64\%, but doesn't work well for the dense attack. It doesn't achieve any improvement for C3D (from 59\% to 59\%). This is expected because the temporal defense makes use of the image patch of
the clean frames to reconstruct the adversarial frames. As for the sparse attack method, there are a
lot of clean frames in the video, this method can use them to compress and reconstruct the
adversarial frames. But as for the dense attack method, all the frames of the video are
attacked, therefore very little available information can be used. Even though
the adversarial frames are compressed and reconstructed, the adversarial
perturbations are still alive.  The results in Table \ref{table:crossstreetmAPresult2} show that temporal defense is more suitable for defending the sparse attack than the dense attack. In addition, we also find that temporal defense is friendly to the clean videos. Both for CNN+LSTM and C3D, the accuracy drop of temporal defense is very slight, and far below than the drop of defense attack. Because of this advantage, we don't need to strictly distinguish the sparse attacks from the clean videos. The temporal defense video frames, and the corresponding
sparse adversarial video frames and the corresponding clean videos are shown in Figure \ref{fig:defense}.

\subsubsection{Video classification with the spatial defense}
The spatial defense is carried out on the same experimental setup as the temporal defense.
In Table \ref{table:crossstreetmAPresult2}, it is clear that
the spatial defense achieves the comparable performance both for sparse attack and dense attack. For the sparse attack, it improves the accuracy of CNN+LSTM from 31\% to 55\%. For the dense attack,
it improves the accuracy of C3D from 59\% to 73\%.
The results show that spatial defense is more suitable for defending the dense attack
method than the temporal attack ( 73\%-59\%=14\% vs 59\%-59\%=0\%).  This is also reasonable because in the dense attack, all the frames are polluted.
In this situation, we can regard the polluted frames as the adversarial images, and directly
apply the image defense method. ComDefend achieves the state-of-the-art image defense performance,
and is also very efficient. Therefore, we choose ComDefend to defend the dense attack.

\subsubsection{Video classification with integrating detection and different defense strategies}
From the above discussions, it is clear that the spatial defense is more suitable to defend
the dense attack method and the temporal defense is more suitable to defend
the sparse attack method.  In order to integrate the advantages of both methods,
we combine the different defense methods with the detection to jointly improve the robustness of video classifiers. According to the result of the detection,
the corresponding defensive method is used to protect the well-trained video classification
model. In particular, as shown in Figure \ref{fig:iccv01}, if the detection result is the clean video,
the original video is fed to well-trained classifier directly. If the detection result is
the sparse adversarial video, the temporal defense is used to deal with the original video. If
the detection result is the dense adversarial video, the spatial dense is used to deal with
the original video. We choose 500 clean videos, 500 temporal adversarial videos and 500
dense adversarial videos to verify our proposed framework. Note that these adversarial
videos can make the well-trained classifier give the wrong label.
As shown in Table \ref{table:crossstreetmAPresult3}, we can see that the spatial and
temporal defense can improve the classification accuracy of the well-trained classifier. Integrating with these two defense methods and the adversarial detection
obtains the best classification accuracy.

\subsection{Result analysis}
In \cite{guo2017countering}, it has demonstrated that the adversarial perturbation of
adversarial example has a particular structure. As for the spatial defense
(ComDefend \cite{jia2018comdefend}), it regards the adversarial perturbation
as a kind of image redundancy information. And then it removes the
adversarial perturbations by image compression. This method is also
a kind of adversarial denoiser. Therefore it is suitable to process the dense adversarial
videos. As for the temporal defense, it makes use of the motion estimation method
to purify the adversarial frames by using the information of clean frames. In particular,
the adversarial frames are reconstructed by the clean frames, motion vectors and
corresponding residual. It destroys the particular structure of the adversarial
perturbations by quantifying the corresponding residual. Because it should
use the information of the clean frames of adversarial video, it is suitable
for the defense for the sparse adversarial videos. We can use the detection, which
is achieved by the temporal consistency between adjacent frames, to
combine the advantages of both defenses to improve model robustness.
That is to say, the selection of the defensive method is dependent on
the result of the detection.

\section{Conclusion}
In this paper, we propose an effective defense framework to characterize
and defend adversarial videos. We are the first one to explore
the defense for adversarial videos. In light of the property of
the state-of-the-art attacking methods, we first apply the detection network to
detect whether the input video is attacked and determine the type of
attack method. And then we select the corresponding defensive methods according to
the result of the detection. A series of experiments demonstrates that
the proposed method can greatly improve the accuracy of the well-trained
video classifier and defend the state-of-the-art attack methods
for video classification.

\small
\bibliographystyle{aaai}
\bibliography{Bibliography-File}

\end{document}